\documentclass[letterpaper]{article} 
\usepackage{aaai24}  
\usepackage{times}  
\usepackage{helvet}  
\usepackage{courier}  
\usepackage[hyphens]{url}  
\usepackage{graphicx} 
\usepackage{natbib}  
\usepackage{caption} 
\usepackage{algorithm}
\usepackage{algorithmic}

\usepackage{newfloat}
\usepackage{listings}
\DeclareCaptionStyle{ruled}{labelfont=normalfont,labelsep=colon,strut=off} 
\floatstyle{ruled}
\newfloat{listing}{tb}{lst}{}
\floatname{listing}{Listing}
\usepackage{latexsym}
\usepackage{booktabs}
\usepackage{xcolor}
\usepackage{color}
\definecolor{cite}{HTML}{D9E1F2}
\definecolor{sem}{HTML}{EF8D7C}
\definecolor{add}{HTML}{E4EEDC}
\definecolor{cla}{HTML}{F5C242}

\title{WikiSQE: A Large-Scale Dataset for Sentence Quality Estimation in Wikipedia}
\author{
  }
\author {
    Kenichiro Ando\textsuperscript{\rm 1},
    Satoshi Sekine\textsuperscript{\rm 1},
    Mamoru Komachi\textsuperscript{\rm 2}
}
\affiliations {
    \textsuperscript{\rm 1}RIKEN AIP
    \textsuperscript{\rm 2}Hitotsubashi University\\
    kenichiro.ando@riken.jp
}

\begin{document}

\maketitle

\begin{abstract}
Wikipedia can be edited by anyone and thus contains various quality sentences.
Therefore, Wikipedia includes some poor-quality edits, which are often marked up by other editors.
While editors' reviews enhance the credibility of Wikipedia, it is hard to check all edited text.
Assisting in this process is very important, but a large and comprehensive dataset for studying it does not currently exist.
Here, we propose WikiSQE, the first large-scale dataset for sentence quality estimation in Wikipedia.
Each sentence is extracted from the entire revision history of English Wikipedia, and the target quality labels were carefully investigated and selected.
WikiSQE has about 3.4 M sentences with 153 quality labels.
In the experiment with automatic classification using competitive machine learning models, sentences that had problems with citation, syntax/semantics, or propositions were found to be more difficult to detect.
In addition, by performing human annotation, we found that the model we developed performed better than the crowdsourced workers.
WikiSQE is expected to be a valuable resource for other tasks in NLP.
\end{abstract}

\section{Introduction}
Wikipedia is a huge online encyclopedia that is famously editable by any user.
It contains various topics and continues to improve its quality through repeated editing by users.
However, the quality of Wikipedia has long been a matter of dispute \cite{Giles2005,britannica2006,Editorial2006,Chesney2006} and is a very important issue for natural language processing (NLP).
Due to Wikipedia texts being widely used in NLP datasets and being a major source \cite{rajpurkar2016,thorne2018,Koupaee2018,banon2020}, the impact of Wikipedia quality on NLP is significant. 

In fact, some poor edits exist caused by abuse, etc., which are often corrected by other editors.
However, checking and revising all poor edits is unrealistic, and there have been several previous attempts to support this problem by machine.
A typical one is Wikipedia's Bot \footnote{\url{https://en.wikipedia.org/wiki/Wikipedia:Bots}}, which automatically corrects errant markup formats, substitutes deprecated features, and, in particular, reverts vandalistic edits.
However, they are targeted at naive and superficial errors, while the remaining more multi-dimensional and in-depth evaluations are marked up by humans.
Other work includes attempting to automatically detect specific labels given by editors \cite{ganter2009,redi2019,bertsch2021}.
The former is unable to make fine-grained evaluations of each sentence, while the latter focuses only on one particular poor-quality aspect.

Hence, we built WikiSQE, a dataset to estimate the quality of sentences in fine-grained and various aspects.
WikiSQE enables the evaluation of various quality aspects that could not be assessed before, including grammatical errors, semantic weirdness, the need for additional information, and many other aspects.
Sentences were acquired from the entire revision history of English Wikipedia.
They are assigned Wikipedia's inline template labels \footnote{\url{https://en.wikipedia.org/wiki/Wikipedia:WikiProject\_Inline\_Templates}} by the Wikipedia editors as sentence quality.
We carefully selected the target labels manually and filtered out noisy sentences, resulting in a total of 153 quality labels and about 3.4 M sentences.
The 153 quality labels were further organized into five categories that we have defined.

\begin{table*}[t!]
    \centering
    \begin{tabular}{lrl}
    \toprule
    \textbf{Group} & \textbf{Count} & \textbf{Description}\\
    \midrule
    \textbf{Citation} &  & \\
	\ \ \ \ Citation needed & 2,373,911 & Citation is required to verify the content.\\
    \ \ \ \ Dead link & 84,101 & External link is broken.\\
    \ \ \ \ Original research? & 69,449 & Cited source is not verified by a third party.\\
    \ \ \ \ Not in citation given & 35,278 & Failed to verify the statement's content from the source.\\
    \ \ \ \ Unreliable source?  & 25369 & The editor cannot trust the source.\\
    \textbf{Syntactic or semantic revision} &  &\\
    \ \ \ \ Clarification needed & 138,739 & Statement is difficult to understand.\\
    \ \ \ \ Vague  & 13,373 & Contains vague words or statement.\\
    \ \ \ \ Check quotation syntax  & 1,272 & Quotation syntax is not match the guidelines.\\
    \ \ \ \ Weasel words & 1,055 & Contains weasel words.\\
    \ \ \ \ Jargon & 636 & Overly jargonistic and too technical statement.\\
    \textbf{Information addition} &  &\\
    \ \ \ \ Who? & 91,924 & Contains claims that do not identify individuals.\\
    \ \ \ \ When? & 72,920 & Time period is so vague or ambiguous.\\
    \ \ \ \ By whom? & 41,588 & Contains a vague, third-party claim that do not identify individuals.\\
    \ \ \ \ Pronunciation? & 31,517 & Pronunciation by audio is needed.\\
    \ \ \ \ Which?  & 23,387 & References to organizations or other things are vague.\\
    \textbf{Disputed claim} &  &\\
    \ \ \ \ Dubious & 45,920 & Sourced statement, but that seems dubious or unlikely.\\
    \ \ \ \ Neutrality disputed  & 8,465 & Statement seemed to be biased.\\
    \ \ \ \ Undue weight?  & 3,544 & Undue weight to an idea or point of view.\\
    \ \ \ \ relevant? & 3,494 & Uncertain if a statement is relevant to the article, or encyclopedic.\\
    \ \ \ \ Disputed  & 2,433 & Statement whose truth or factual is in dispute by editors.\\
    \textbf{Other} &  &\\
	\ \ \ \ Disambiguation needed & 107,953 & Contains a wikilink which should be linked to a specific page.\\
    \ \ \ \ Sic & 50,658 & Textual error in the statement is copied exactly from the source.\\
    \ \ \ \ Needs update  & 19,550 & Statement needs to be updated based on recent events.\\
    \ \ \ \ Specify  & 7,069 & Sourced statement, but not sure of alignment with sources. \\
    \ \ \ \ Emphasis added  & 2,444 & New emphasis added to the quotation.\\
    \midrule
    \textbf{Total} & 3,417,955 & \\
    \bottomrule
    \end{tabular}
    \caption{\label{table:desc}
    Examples of quality labels in Wikipedia that are classified into five categories.
    Five interesting labels are selected from the top ten most frequent labels in each category, with the total number of sentences and their description.
    }
\end{table*}

The experiments with automatic classification using competitive machine learning models found that statements requiring additional information are classified with high accuracy, but that sentences requiring resolving citation, syntactic or semantic revision, or propositional problems are difficult to detect.
In the human annotation experiment using crowdsourcing, we found that machine learning models we developed performed better than the crowdsourced workers.

Besides, our dataset is expected to be a valuable resource for other tasks in NLP.
The editor's quality-checking process can be seen as also a professional annotation.
They have experience in the verification of many expressions and left the results of their sentence quality assessments on Wikipedia's labels.
In particular, given the growing importance of corpus filtering in training large language models to improve the quality of the corpus used, this dataset is very valuable.
The dataset and codes used in this work are publicly available \footnote{\url{https://github.com/ken-ando/WikiSQE}}.

\section{Related Work}
Previous studies on \textbf{quality estimation in Wikipedia} have mainly focused on the article-level \cite{Mola2011,Bykau2015,wong2021,Asthana2021}.
They are aimed at estimating the quality of revisions and articles.
On the other hand, quality estimation studies focusing on sentence-level also exist.
The quality labels for detection were created by manually assigned to small-scale sentences \cite{herzig2011,Hube2019} or by using Wikipedia's Inline template.
For the latter, Inline templates include ``Citation needed'' \cite{redi2019}, which indicates that a sentence needs citations, ``Puffery'' and ``Peacock'' \cite{bertsch2021}, which indicates that the sentence contains exaggerated expressions, and the ``Weasel words'' \cite{ganter2009}, which indicates that the sentence contains ambiguous wording.
A detailed comparison of our study with previous studies is given in Table \ref{tab:comp}.
Previous studies cover a subset of Wikipedia quality labels and do not include various labels.
Regarding the availability of data, two datasets are not publicly available.
In comparison, our dataset contains a large number of labels and sentences and is available for public access.

\begin{table}[t]
\centering
\begin{tabular}{lrc}
\toprule
\textbf{Inline templete} & \textbf{\# Sents} & \textbf{Available?} \\
\midrule
Citation needed & 36,140 & No \\
Weasel words & 500 & No\\
Peacock, Puffery & 284 & Yes\\
\bottomrule
\end{tabular}
\caption{\label{tab:comp}Comparison with previous studies. ``Available?'' indicates the public availability of the data.}
\end{table}

Similar tasks for \textbf{sentence quality estimation in other domains} include grammatical error correction \cite{ng2014}, linguistic acceptability \cite{warstadt2019}, automated essay estimation \footnote{\url{https://www.kaggle.com/c/asap-aes}}, etc.
These are related to our sentence quality estimation task on Wikipedia.
Indeed, our dataset contains labels that point out grammatical errors or semantically weird expressions (see Section ``Analysis of WikiSQE'').
Therefore, it may be possible to help with these related tasks.
Others, such as ``Neutrality disputed'' is expected to be a biased sentence set, so it is relevant to the existing bias detection dataset\cite{stereoset} and may be useful for the bias detection and removal task.
Similarly, ``Dubious'' is a factually dubious sentence set, so it is relevant to the existing fact-checking dataset\cite{thorne-etal-2018-fever} and could be useful.

\section{Sentence Quality Estimation Dataset}

\subsection{Source Text}
Wikipedia is written in a markup language called Wiki markup, and the officially provided dump file of Wikipedia \footnote{\url{https://dumps.wikimedia.org/enwiki/}} is also written in Wiki markup.
To extract sentences for our dataset, we need to convert this dump file to HTML.
The most accurate parser is the official MediaWiki \footnote{\url{https://www.mediawiki.org/wiki/MediaWiki}}, which is also computationally expensive.
This study targets the entire edit history of the English version, which requires especially large computational resources and time.
Fast-processing third-party HTML conversion tools can be used, but WikiMedia's frequent version upgrades can cause many such tools to fail.
For this reason, we use the full edit history data that has already been converted to HTML using MediaWiki in the previous study  \cite{mitrevski2020}\footnote{Creative Commons Attribution 3.0 Unported}.
This data contains the entire edit history of all articles on English Wikipedia before 1 March 2019.
A total of 7 TB of files need to be downloaded to use it.

\subsection{Quality Label}\label{target}
The target quality labels for constructing our dataset are those contained in Wikipedia's inline cleanup template \footnote{\url{https://en.wikipedia.org/wiki/Category:Inline_cleanup_templates}}, 344 templates included in all.
This collection of inline templates is a set of Wiki markups for editors to point out the poor quality of sentences in Wikipedia.
This set includes some templates that are not related to sentence quality, such as those used for talk pages and user pages.
Therefore, we carefully hand-picked 147 templates from 344.
Next, these inline templates need to be converted to HTML to match the source text.
Therefore, we converted these inline templates to HTML using the Wikipedia sandbox, thus obtaining an initial list of quality labels.

However, there are still some problems regarding the coverage of our quality label list.
The first problem is the temporal differences in the MediaWiki parser.
The HTML of the source text was generated by MediaWiki as of the year 2019, thus there is a difference from the HTML output by the current MediaWiki.
This means that HTML labels acquired by the current sandbox may not be present in the source text.
To address this problem, we manually checked high-frequency inline templates in the source text and added new ones to our quality label list.
As a result, six new labels were added, yielding a total of 153 quality labels.
Some of them are shown in Table \ref{table:desc}.
All obtained quality labels and their descriptions are available on the web page of WikiSQE.
Second, there are temporal differences in the Wikipedia inline cleanup template.
Our Wikipedia inline cleanup template used for the quality labels is from the year 2022, which differs from the 2019 version existing in the source text.
Hence, we need to obtain the inline template in the year 2019.
We acquired the past quality labels by recursively getting the already deprecated inline template pages that redirect to each inline template page.
As a result, 1,319 past inline templates linked to 153 quality labels were acquired.
In addition, for all 153 labels, we manually matched past and current labels and linked them to the corresponding inline template description page in Wikipedia.
The information is posted on the WikiSQE page.
Therefore, even if inline template names change in the future, they can be uniquely identified.

\begin{table*}[t!]
    \centering
    \begin{tabular}{p{3.8cm}p{11.3cm}}
    \toprule
    \textbf{Label} & \textbf{Sentence} \\
    \midrule
	Citation needed & According to Japanese records, the term kendo is coined in Japan on August 1, 1919.[citation needed]\\
    Dead link & Player profile at LFChistory.net[dead link]\\
	Clarification needed & It was later given to the county, and has a possibility of becoming County Road 23.[clarification needed]\\
    Vague & Pisces is perhaps[vague] the first hit rock or pop album to feature the Moog.\\
	Who? & However, many analysts[who?] are finding that as Google grows, the company is becoming more "corporate".\\
    When? & Over the last thirty years,[when?] a debate has been ongoing whether a tiny number of Ukrainians settled in Canada before 1891.\\
    Disambiguation needed & Attala County, Mississippi: Attala is named for Attala [disambiguation needed], a fictional Native American heroine.\\
	Sic & He also notably sung 'Digital Surviver[sic]', theme of Akiyama Ryo from Digimon Tamers.\\
	Dubious & Some academic linguists believe the modern English Language is half-Romance influenced (the evident Norman French influences), thus can be classified a Romance language.[dubious]\\
    Neutrality disputed & Streetball is a very popular game worldwide, and a fun way for young people to keep out of trouble and avoid problems such as juvenile crime and drugs.[neutrality disputed]\\
    \bottomrule
    \end{tabular}
    \caption{\label{exsamples}
    Examples of quality labels in Wikipedia and their sentences.
    }
\end{table*}

\subsection{Quality Category}
For the analysis, we categorized the 153 quality labels into five types that were further abstracted by their characteristics (Table \ref{table:desc}, Table \ref{statics}).
The five categories we defined and their descriptions are as follows.

\textbf{Citation} category contains 59 labels related to the citation and is the most sentence-rich category.
They include mentions that some citation is needed, the quality of the reference, problems with the format of the citation, and problems with the sentence cannot be reconstructed from the reference.
The majority of the label is ``Citation needed'', which indicates that the sentence needs some citation, and it accounts for 69\% of the labels across the entire dataset.

\textbf{Syntactic or semantic revision} is the category that indicates that grammatical or semantic improvement is needed, to which 26 labels belong.
``Clarification needed'' is the most common label indicating that the unclear and difficult-to-understand meaning of the text should be clarified.
There is also a label ``Check quotation syntax'', which indicates that the format of the quotation is incorrect.

\textbf{Information addition} is a category that points out the need for some additional information in the sentence.
The most common label is ``Who?'', which indicates that the name of a specific person or organization is not specified.
Similarly, there are many labels that require other entities, such as locations or times, to be specified.

\begin{table*}[t!]
    \centering
    \begin{tabular}{lrrrr}
    \toprule
    \textbf{Group} & \textbf{Count} & \textbf{\# Labels} & \textbf{Ave. \# Tokens} & \textbf{Perplexity} \\
    \midrule
    Citation & 2,687,535 & 59 & 26.83 & 90.48\\
    Syntactic or semantic revision& 160,350 & 26 & 27.01 & 110.53\\
    Information addition& 310,853 & 32 & 27.82 & 79.37\\
    Disputed claim & 70,202 & 20 & 28.30 & 82.22\\
    Other & 188,969 & 16 & 34.01 & 110.92\\
    \bottomrule
    \end{tabular}
    \caption{\label{statics}
    Statistics of five sentence quality categories.
    \# Labels denotes the number of labels, \# Tokens denotes the average number of words in sentences, and Perplexity denotes the average perplexity of sentences.
    }
\end{table*}

\textbf{Disputed claim} is a category indicating that there is no problem with the form of the sentence, but that there is a problem with its proposition.
The most common label is Dubious, which indicates that information is unnatural and of dubious truth from the editors.
This category also includes statements that are not neutral, have some kind of bias, or are not suitable for an encyclopedia.

Finally, \textbf{Other} is a set of labels that do not belong to any other category.
The most common label is ``Disambiguation needed'', which is marked up when the Wikilink in the sentence is linked to a disambiguation page and needs to be improved.
The next most common label is ``Sic''.
This label contains sentences that are faithful to references but are generally grammatically incorrect.

These categories are further classified as Wikipedia-dependent and Wikipedia-independent.
The three categories of Syntactic or semantic revision, Information addition, and Disputed claim have characteristics that can be generalized to common sentence quality estimation.
Therefore, they are classified as \textbf{Common} and handled separately in later experiments.
``Common'' contains 78 different labels, which is about half of all labels.

\subsection{Sentence Extraction and Filtering}\label{filtering}
To extract sentences from the source text, we first split sentences using pySBD \cite{sadvilkar2020} and then extracted sentences whose label was included in the quality label list obtained in Section \ref{target}.
However, the extracted sentences are noisy as they often contain items such as section titles and non-sentences.
Therefore, we filtered out extremely short sentences of less than 10 words, sentences with the Wiki markup, and sentences  having lower-case initial letters.
Each sentence was stripped of citation markers and quality labels, and duplicates were removed, resulting in a final set of 3,417,909 sentences.
For later experiments, all quality labels are removed from the sentences here, but the position of the quality labels is very important information, so our public dataset includes the version in that sentences have quality labels.

\begin{table*}[t!]
    \centering
    \begin{tabular}{lccc}
    \toprule
    \textbf{Group} & \textbf{DeBERTa} & \textbf{BERT} & \textbf{RoBERTa} \\
    \midrule
    Citation & 73.0 & \textbf{74.5} & 73.9\\
    Syntactic or semantic revision& 73.0 & \textbf{73.8} & 71.9\\
    Information addition& 85.2 & \textbf{85.3} & 83.3\\
    Disputed claim & \textbf{74.3} & 73.2 & 74.2\\
    Other & \textbf{82.6} & 80.6 & 81.3\\
    \midrule
    Common & 79.5 & 79.0 & \textbf{80.0}\\
    All & 70.4 & \textbf{72.3} & 71.6\\
    \bottomrule
    \end{tabular}
    \caption{\label{res_cat}
    Experimental results of automatic detection for each quality category.
    \textbf{Bolded text} indicates the best F1-scored model.
    ``Common'' is a set of ``Syntactic or semantic revision'', ``Information addition'', and ``Disputed claim''.
    }
\end{table*}

\begin{table}[t!]
    \centering
     \begin{tabular}{lrc}
    \toprule
    \textbf{Label} & \textbf{Count} & \textbf{F1} \\
    \midrule
    \colorbox{cite}{Citation needed}&2,373,911 & 70.0\\
    \colorbox{sem}{Clarification needed}&138,739 & 74.8\\
    \colorbox{cite}{Disambiguation needed}&107,953 & 82.3\\
    \colorbox{add}{Who?}&91,924 & 88.7\\
    \colorbox{cite}{Dead link}&84,101 & 77.9\\
    \colorbox{add}{When?}&72,920 & 87.4\\
    \colorbox{cite}{Original research?}&69,449 & 89.9\\
    \colorbox{white}{Sic}&50,658 & 92.9\\
    \colorbox{cla}{Dubious}&45,920 & 75.6\\
    \colorbox{add}{By whom?}&41,588 & 92.9\\
    \colorbox{cite}{Not in citation given}&35,278 & 75.2\\
    \colorbox{add}{Pronunciation?}&31,517 & 98.8\\
    \colorbox{cite}{Attribution needed}&27,322 & 94.3\\
    \colorbox{cite}{Unreliable source?}&25,369 & 74.1\\
    \colorbox{add}{Which?}&23,387 & 86.0\\
    \colorbox{white}{Needs update}&19,550 & 92.0\\
    \colorbox{cite}{Verification needed}&18,311 & 74.2\\
    \colorbox{add}{According to whom?}&15,205 & 83.1\\
    \colorbox{sem}{Vague}&13,373 & 74.1\\
    \colorbox{cla}{Neutrality disputed}&8,465 & 82.7\\
    \bottomrule
    \end{tabular}
    \caption{\label{res_each}
    Results of automatic detection of the top 20 quality labels in frequency using DeBERTa. Each highlight represents the quality category of \colorbox{cite}{Citaion}, \colorbox{sem}{Syntactic or semantic revision}, \colorbox{add}{Information addition}, \colorbox{cla}{Disputed claim}, or \colorbox{white}{Other}.
    }
\end{table}

\subsection{Analysis of WikiSQE}\label{ana_wikisqe}
Examples of sentences included in WikiSQE are shown in Table \ref{exsamples}.
We found that some labels are added to sentences, words, and clauses.
For example, ``Citation needed'', ``Clarification needed'', ``Dubious'', and ``Neutrality disputed'' are, by their characteristics, frequently added to specific spans or clauses in sentences.
On the other hand, ``Who?'', ``When?'' and ``Sic'' are often added to words.
This indicates that each label needs a different scope when assessing quality; the former cannot be judged without considering the meaning of the entire sentence, while the latter can be judged by considering the words alone.
However, note that "Who?", "When?", etc. can be used to ambiguously request additional information about the whole sentence, so it is not an easy task.

In the sentence example, ``Citation needed'' refers to the date when the term kendo was created and points out that it needs to be supplemented by external documentation.
This is a typical example of ``Citation needed'', which is frequently used to enhance credibility by requesting supporting documentation when a specific date, time, or place is mentioned.
In the example of  ``Clarification needed'', it seems that the editor requests the specific process by which a road becomes County Road 23.
This is a point that cuts deeply into the semantics of the sentence, and ``Clarification needed'' contains many examples that would be difficult to label without linguistic proficiency.
However, these points are very important in making the article more readable.
In ``Who?'' and ``When?'', the editor requests that additional new information in the phrases referring to an ambiguous person and time, respectively.
These two examples are typical, and important regarding the credibility of the sentence gained by clarifying the person and time.
This is especially important when the ambiguous phrase is crucial to the credibility of the sentence.
In the example of ``Sic'', the title of a song is labeled as an official name, albeit grammatically incorrect.
``Sic'' is most commonly assigned to such proper nouns, and quotation marks are often used to state that the notation is correct.
In ``Dubious'', the editors take issue with the claim that English is classified as the Romance language.
It is very difficult to judge such claims, as they require a high level of intelligent work.
It is necessary to evaluate in reference to one's own common sense and knowledge, whose process is closely related to fact-checking.
In ``Neutrality disputed'', the editors label statements that seem to overstate the positive impact of street basketball.
Since Wikipedia is an encyclopedia, biased statements are not appropriate.
However, editors often disagree about what expressions are biased.
Therefore, discussions are usually held on the talk page.

Table \ref{statics} shows the perplexity of each category calculated using GPT-2 \cite{radford2019}.
Each perplexity reflects the characteristics of the category well, with ``Syntactic or semantic revision'' having a high perplexity because it contains many sentences with usage that is grammatically and semantically unusual.
For example, ``Jargon'' includes rare technical terms.
``Other'' has particularly high perplexity because it contains many sentences with ``Sic''.
``Sic'' includes misspellings, which are major factors in raising perplexity.
``Disputed claim'' has low perplexity, although it is propositionally unusual, so it may be predicted harder because its weirdness does not appear on the surface.
``Information addition'' requires additional information, but the included sentences have no problem, and thus have low perplexity.

\section{Experiment}
We performed experiments to automatically detect problematic sentences in Wikipedia using WikiSQE.

\subsection{Dataset}\label{dataset_class}
To ensure enough size for the development and test data, we performed the experiments by each quality category and by the top 20 most frequent labels.
For the experiments, we use the sentences included in the WikiSQE as positive examples, and for the negative examples, we extract newly unlabeled sentences using the same steps as in section ``Sentence Extraction and Filtering'', and randomly sample the data to the same size as the positive examples.
For the development and test data, we randomly extract 500 positive and 500 negative examples each and concatenate them to make 1000 sentences.
The remaining data were used as training data.

``Citation needed'' contains a significantly large number of sentences, so we downsampled the data to 200,000 sentences in order to evaluate the categories as fairly as possible with respect to other labels.
In addition, we prepare ``All'' category that includes all quality labels.
It is not just a weighted average, but an independent dataset in which the sentences in all categories are concatenated and shuffled.

\subsection{Setup}
The models used for detection are competitive large-scale pre-training models DeBERTaV3 \cite{debertav3} and BERT \cite{devlin2018}, RoBERTa \cite{roberta}.
DeBERTa and RoBERTa both used base models, and BERT used a base\_uncased model.
Fine-tuning was performed with the method normally used for classification tasks in NLP.
The multi-layer perceptron is used as the final layer and trained to classify in two classes, positive and negative, whether a sentence belongs to a target quality label or not.
The maximum number of training epochs is 20, and the model that records the highest F1 in the development set is used as the best model to predict the test set.
The learning rate is determined by searching among 1-e6, 5-e6, 1-e5, and 5-e5.
The maximum input sequence length is 256 and the batch size is 64.
In all setups, we report the average F1 values of the experiments with three different seeds.

\subsection{Results}
The experimental results of the automatic detection for the categories are shown in Table \ref{res_cat}.
The overall F1 value was 70--85\%, but detection of ``Citation'', ``Syntactic or semantic revision'', and ``Disputed claim'' were relatively difficult.
``Citation'' seems to be difficult to identify using Wikipedia alone, since it must consider the contents of the references.
For example, ``Original research?'' and ``Unreliable source?'' depend on the reliability of the cited documents.
``Syntactic or semantic revision'' and ``Disputed claim'' are also challenging because they require capturing higher-level semantic aspects.
They are less than other categories to have features such as numbers or names of people appear on the surface of the sentence.
``Information addition'' was detected with high accuracy because the expressions to which the information is added are unique (e.g., When? is for temporal expressions).

Comparisons among the models showed that BERT had the best overall score, but there were no significant differences in performance.
``All'' category provided the lowest performance for any of the models.
On the other hand, ``Common'' had higher performance, and considering the difference, this may be the result of training on cross-category data, ignoring the clustering of categories.
``Common'' cluster similar sentences better than ``All'', and thus seems to have learned well.

The results of the automatic detection experiment for individual quality labels are shown in Table \ref{res_each}.
It can be observed that the F1 score changes a lot depending on the label.
Those with F1 scores above 90 are ``Sic'', ``Pronunciation?'', ``Attribution needed'', and ``Needs update'', which are relatively easy to detect.
The reasons are that ``Sic'' is characterized by grammatical errors and ``Pronunciation?'' is characterized by frequent sentences containing other languages.
``Attribution needed'' is a label used to request attribution when quoting someone's statement, etc., and is unique.
``Needs update'' is used mainly in numerical values of sports articles and is easy to detect.
Thus, even for quality labels in the same category, detection performance differs greatly depending on the characteristics of the label.

\begin{table}[t!]
    \centering
     \begin{tabular}{lccc}
    \toprule
    \textbf{Label} & \textbf{Human} & \textbf{DeBERTa} & \textbf{GPT-4}\\
    \midrule
    \colorbox{cite}{Citation needed}& 75.5 & 69.5 & 67.0\\
    \colorbox{sem}{Clarification needed}& 61.0 & 74.0 & 66.0\\
    \colorbox{add}{Who?}&69.1 & 88.4 & 81.0\\
    \colorbox{add}{When?}&71.8 & 87.0 & 64.0\\
    \colorbox{white}{Sic}&73.4 & 92.7 & 83.0\\
    \colorbox{cla}{Dubious}&65.5 & 74.0 & 73.0\\
    \colorbox{add}{Pronunciation?}&70.9 & 98.7 & 95.0\\
    \colorbox{add}{Which?}&75.5 & 85.7 & 62.0\\
    \colorbox{white}{Needs update}&76.4 & 92.0 & 85.0\\
    \colorbox{sem}{Vague}&64.5 & 73.5 & 66.0\\
    \colorbox{cla}{Neutrality disputed}&77.3 & 81.8 & 83.0\\
    \bottomrule
    \end{tabular}
    \caption{\label{res_annotation}
    Results of cloud-based annotation.
``Human'' shows the average accuracy of annotators' answers.
``DeBERTa'' shows the average accuracy of the same models in Table \ref{res_each}. ``GPT-4'' is the result of using GPT-4 with 4-shot.
    }
\end{table}

\section{Dataset Analysis via Non-expert Annotation}\label{annotation}
To more deeply investigate the quality and potential of our dataset, we conducted annotation experiments.
For this annotation, we used the crowdsourcing platform APPEN, so almost all annotators were not experts on Wikipedia, like editors.
It is difficult to accurately assign the various Wikipedia inline templates without well-trained editors, so this experiment can be considered as a comparison between the performance of our model trained on WikiSQE and that of non-experts.
The labels for annotation were selected from the 20 quality labels used in the model experiments due to cost constraints, excluding the set of labels that could not be answered correctly without understanding the external source.

The annotation procedure is the following pairwise method.
First, we extracted one sentence each from the set of sentences with target quality labels and the set of sentences without quality labels.
Second, both were presented to annotators with label descriptions and examples.
For the explanation of the quality labels, we referred to Wikipedia's Inline template pages and wrote as clearly and in as much detail as possible.
Finally, workers judged which sentence was more appropriate for the target quality label.
We created 10 pairs of sentences per quality label for all 11 quality labels and assigned 10 annotators to each pair.
To screen out good workers, easy examples were selected from the dataset and tested at the beginning of the annotation.
Annotators are limited to U.S. residents who select English as their language of use in the Appen platform.

The annotation results are shown in Table \ref{res_annotation}.
``Neutrality disputed'', ``Needs update'', ``Which?'', and ``Citation needed'' indicate relatively good accuracy, indicating that these labels are easily recognizable by humans.
Conversely, ``Clarification needed'' has low accuracy and seems to be hard to pick sentences that need especially clarification.
By category, we found that ``Syntactic or semantic revision'', which require a high level of semantic handling, are all low, but ``Other'', with its superficial features, is relatively high.
Compared to the model's performance, we found that the model is better for all labels except ``Citation needed''.
This indicates that the model outperformed non-experts unfamiliar with editing Wikipedia by learning from expert-generated data.
This result confirms the usefulness of the WikiSQE.

We also experimented using GPT-4\cite{gpt4} with the 4-shot setting.
As a prompt, we wrote a short description of a target label and an instruction that let the GPT-4 choose which of two sentences is more appropriate for the target label.
Four examples and their answers were also added to prompts as in-context learning.
In the experiment, we randomly changed the four examples and evaluated 100 sentence pairs for each quality label.
We put ``You are an excellent assistant to Wikipedia editors.'' as the system role, and added "Description:", "Examples:", and "Question:" at the head of each label description, few-shot examples, and target two choices, respectively. 
The results showed the lowest scores for 3 of 11 labels, and 10 of 11 labels lost to DeBERTa.
Compared to the fine-tuned model, we found that even with the GPT-4, many labels failed to be answered with only 4 shots and a label description.
However, even without fine-tuning, we found that many labels were better classified with GPT-4 than with non-expert annotations.
This indicates that GPT-4 is superior in identifying biased or prejudiced expressions, which may be the result of GPT-4's careful training for safety.

\section{Conclusion and Future Work}
In this study, we constructed a sentence quality estimation dataset for Wikipedia.
We obtained a total of 153 quality labels using Wikipedia's inline template and applied them to the entire edit history of English Wikipedia, resulting in 3,417,955 sentences.
Sentences were labeled with a variety of quality aspects, and we further classified them into five upper categories.
In automatic detection experiments for coarse-grained categories, we found that the model was able to detect poor-quality sentences with an F1 score of 70--85\%.
We also found that ``Citation'' category, which is difficult to detect based on Wikipedia articles alone, and ``Syntactic or semantic revision'' and ``Disputed claim'' categories, which require a high level of semantic interpretation for the model, are relatively difficult to detect.
In the automatic detection experiments for the fine-grained labels, we found that the detection performance varied greatly depending on the labels' characteristics.
In human annotation, we demonstrated the superiority of the model we developed by outperforming the accuracy of crowd workers.

We have decided not to make it multilingual because we had to spend a lot of time just preparing the English version, but it could create multilingual versions using the methodology in this paper.
They should be as useful as the English version.
In addition, several application studies using WikiSQE were possible, but we did not include them in this paper.
These are expected to be conducted in the future.

\section{Limitations}
Our limitations are mainly regarding test data.
We did not evaluate detection performance with the realistic distribution on Wikipedia at testing time.
In this paper, we constructed the same amount of positive and negative examples, but in reality, there are quite a few positive examples relative to the negative examples.
For example, the ``Citation needed'' label, which is among the most common, appears in only 0.02\% of all sentences, so the occurrence of other labels is even rarer.
This means that the model must find only a few problematic sentences out of a huge amount of sentences, and naively experimenting in this setting will not draw the characteristics of WikiSQE.
For this reason, we did not experiment with a realistic distribution.
This issue could be mitigated with a pipeline strategy such as pre-filtering.

\section{Ethical Considerations}
This dataset may contain offensive, biased, or discriminatory sentences due to the high volume of user edits.
However, since the purpose of this study is to evaluate sentences comprehensively, including such a perspective, no filtering was performed.

\section{Acknowledgement}
This research was supported by the National Institute of Information and Communications Technology Contract Research Programme 22501 and the JSPS KAKENHI JP20269633.
We are thankful to Mr. Koichiro Watanabe for insightful discussions.

\bibliography{aaai24}

\begin{thebibliography}{28}
\providecommand{\natexlab}[1]{#1}

\bibitem[{Asthana et~al.(2021)Asthana, Tobar~Thommel, Halfaker, and
  Banovic}]{Asthana2021}
Asthana, S.; Tobar~Thommel, S.; Halfaker, A.~L.; and Banovic, N. 2021.
\newblock Automatically Labeling Low Quality Content on Wikipedia By Leveraging
  Patterns in Editing Behaviors.
\newblock \emph{Proceedings of the ACM on Human-Computer Interaction},
  5(CSCW2).

\bibitem[{Ba{\~n}{\'o}n et~al.(2020)Ba{\~n}{\'o}n, Chen, Haddow, Heafield,
  Hoang, Espl{\`a}-Gomis, Forcada, Kamran, Kirefu, Koehn, Ortiz~Rojas,
  Pla~Sempere, Ram{\'\i}rez-S{\'a}nchez, Sarr{\'\i}as, Strelec, Thompson,
  Waites, Wiggins, and Zaragoza}]{banon2020}
Ba{\~n}{\'o}n, M.; Chen, P.; Haddow, B.; Heafield, K.; Hoang, H.;
  Espl{\`a}-Gomis, M.; Forcada, M.~L.; Kamran, A.; Kirefu, F.; Koehn, P.;
  Ortiz~Rojas, S.; Pla~Sempere, L.; Ram{\'\i}rez-S{\'a}nchez, G.; Sarr{\'\i}as,
  E.; Strelec, M.; Thompson, B.; Waites, W.; Wiggins, D.; and Zaragoza, J.
  2020.
\newblock {P}ara{C}rawl: Web-Scale Acquisition of Parallel Corpora.
\newblock In \emph{Proceedings of the 58th Annual Meeting of the Association
  for Computational Linguistics}, 4555--4567.

\bibitem[{Bertsch and Bethard(2021)}]{bertsch2021}
Bertsch, A.; and Bethard, S. 2021.
\newblock Detection of Puffery on the {E}nglish {W}ikipedia.
\newblock In \emph{Proceedings of the Seventh Workshop on Noisy User-generated
  Text}, 329--333.

\bibitem[{Britannica(2006)}]{britannica2006}
Britannica, E. 2006.
\newblock Fatally Flawed: Refuting the Recent Study on Encyclopedic Accuracy by
  the Journal Nature.
\newblock \emph{Chicago, Estados Unidos: Encyclopaedia Britannica}.

\bibitem[{Bykau et~al.(2015)Bykau, Korn, Srivastava, and
  Velegrakis}]{Bykau2015}
Bykau, S.; Korn, F.; Srivastava, D.; and Velegrakis, Y. 2015.
\newblock Fine-grained Controversy Detection in Wikipedia.
\newblock In \emph{2015 IEEE 31st International Conference on Data
  Engineering}, 1573--1584.

\bibitem[{Chesney(2006)}]{Chesney2006}
Chesney, T. 2006.
\newblock An empirical examination of Wikipedia’s credibility.
\newblock \emph{First Monday}, 11(11).

\bibitem[{Devlin et~al.(2019)Devlin, Chang, Lee, and Toutanova}]{devlin2018}
Devlin, J.; Chang, M.-W.; Lee, K.; and Toutanova, K. 2019.
\newblock {BERT: Pre-training of Deep Bidirectional Transformers for Language
  Understanding}.
\newblock In \emph{Proceedings of the 2019 Conference of the North {A}merican
  Chapter of the Association for Computational Linguistics: Human Language
  Technologies}, 4171--4186.

\bibitem[{Editorial(2006)}]{Editorial2006}
Editorial. 2006.
\newblock Britannica Attacks.
\newblock \emph{Nature}, 440: 582.

\bibitem[{Ganter and Strube(2009)}]{ganter2009}
Ganter, V.; and Strube, M. 2009.
\newblock {Finding Hedges by Chasing Weasels: Hedge Detection Using {W}ikipedia
  Tags and Shallow Linguistic Features}.
\newblock In \emph{Proceedings of the 2009 Conference of the North American
  Chapter of the Association for Computational Linguistics: Human Language
  Technologies}, 173--176. Suntec, Singapore.

\bibitem[{Giles(2005)}]{Giles2005}
Giles, J. 2005.
\newblock Internet Encyclopaedias Go Head to Head.
\newblock \emph{Nature}, 438: 900–901.

\bibitem[{He, Gao, and Chen(2021)}]{debertav3}
He, P.; Gao, J.; and Chen, W. 2021.
\newblock DeBERTaV3: Improving DeBERTa using ELECTRA-Style Pre-Training with
  Gradient-Disentangled Embedding Sharing.
\newblock arXiv.

\bibitem[{Herzig, Nunes, and Snir(2011)}]{herzig2011}
Herzig, L.; Nunes, A.; and Snir, B. 2011.
\newblock An Annotation Scheme for Automated Bias Detection in {W}ikipedia.
\newblock In \emph{Proceedings of the 5th Linguistic Annotation Workshop},
  47--55.

\bibitem[{Hube and Fetahu(2019)}]{Hube2019}
Hube, C.; and Fetahu, B. 2019.
\newblock Neural Based Statement Classification for Biased Language.
\newblock In \emph{Proceedings of the Twelfth ACM International Conference on
  Web Search and Data Mining}, 195–203.

\bibitem[{Koupaee and Wang(2018)}]{Koupaee2018}
Koupaee, M.; and Wang, W.~Y. 2018.
\newblock WikiHow: A Large Scale Text Summarization Dataset.
\newblock arXiv preprint.

\bibitem[{Liu et~al.(2019)Liu, Ott, Goyal, Du, Joshi, Chen, Levy, Lewis,
  Zettlemoyer, and Stoyanov}]{roberta}
Liu, Y.; Ott, M.; Goyal, N.; Du, J.; Joshi, M.; Chen, D.; Levy, O.; Lewis, M.;
  Zettlemoyer, L.; and Stoyanov, V. 2019.
\newblock {RoBERTa: A Robustly Optimized BERT Pretraining Approach}.
\newblock arXiv.

\bibitem[{Mitrevski, Piccardi, and West(2020)}]{mitrevski2020}
Mitrevski, B.; Piccardi, T.; and West, R. 2020.
\newblock {WikiHist. html: English Wikipedia's Full Revision History in HTML
  Format}.
\newblock In \emph{Proceedings of the International AAAI Conference on Web and
  Social Media}, volume~14, 878--884.

\bibitem[{Mola-Velasco(2011)}]{Mola2011}
Mola-Velasco, S.~M. 2011.
\newblock Wikipedia Vandalism Detection.
\newblock In \emph{Proceedings of the 28th International Conference on World
  Wide Web Companion}, 391–396.

\bibitem[{Nadeem, Bethke, and Reddy(2021)}]{stereoset}
Nadeem, M.; Bethke, A.; and Reddy, S. 2021.
\newblock {S}tereo{S}et: Measuring Stereotypical Bias in Pretrained Language
  Models.
\newblock In \emph{Proceedings of the 59th Annual Meeting of the Association
  for Computational Linguistics and the 11th International Joint Conference on
  Natural Language Processing}, 5356--5371.

\bibitem[{Ng et~al.(2014)Ng, Wu, Briscoe, Hadiwinoto, Susanto, and
  Bryant}]{ng2014}
Ng, H.~T.; Wu, S.~M.; Briscoe, T.; Hadiwinoto, C.; Susanto, R.~H.; and Bryant,
  C. 2014.
\newblock The {C}o{NLL}-2014 Shared Task on Grammatical Error Correction.
\newblock In \emph{Proceedings of the Eighteenth Conference on Computational
  Natural Language Learning: Shared Task}, 1--14.

\bibitem[{OpenAI(2023)}]{gpt4}
OpenAI. 2023.
\newblock GPT-4 Technical Report.
\newblock arXiv.

\bibitem[{Radford et~al.(2019)Radford, Wu, Child, Luan, Amodei, Sutskever
  et~al.}]{radford2019}
Radford, A.; Wu, J.; Child, R.; Luan, D.; Amodei, D.; Sutskever, I.; et~al.
  2019.
\newblock Language Models are Unsupervised Multitask Learners.
\newblock \emph{OpenAI blog}.

\bibitem[{Rajpurkar et~al.(2016)Rajpurkar, Zhang, Lopyrev, and
  Liang}]{rajpurkar2016}
Rajpurkar, P.; Zhang, J.; Lopyrev, K.; and Liang, P. 2016.
\newblock {SQ}u{AD}: 100,000+ Questions for Machine Comprehension of Text.
\newblock In \emph{Proceedings of the 2016 Conference on Empirical Methods in
  Natural Language Processing}, 2383--2392.

\bibitem[{Redi et~al.(2019)Redi, Fetahu, Morgan, and Taraborelli}]{redi2019}
Redi, M.; Fetahu, B.; Morgan, J.; and Taraborelli, D. 2019.
\newblock Citation Needed: A Taxonomy and Algorithmic Assessment of Wikipedia's
  Verifiability.
\newblock In \emph{Proceedings of the 28th International Conference on World
  Wide Web Companion}, 1567–1578.

\bibitem[{Sadvilkar and Neumann(2020)}]{sadvilkar2020}
Sadvilkar, N.; and Neumann, M. 2020.
\newblock {PySBD: Pragmatic Sentence Boundary Disambiguation}.
\newblock In \emph{Proceedings of Second Workshop for NLP Open Source
  Software}, 110--114. Online.

\bibitem[{Thorne et~al.(2018{\natexlab{a}})Thorne, Vlachos, Christodoulopoulos,
  and Mittal}]{thorne2018}
Thorne, J.; Vlachos, A.; Christodoulopoulos, C.; and Mittal, A.
  2018{\natexlab{a}}.
\newblock {FEVER}: a Large-scale Dataset for Fact Extraction and
  {VER}ification.
\newblock In \emph{Proceedings of the 2018 Conference of the North {A}merican
  Chapter of the Association for Computational Linguistics: Human Language
  Technologies}, 809--819.

\bibitem[{Thorne et~al.(2018{\natexlab{b}})Thorne, Vlachos, Christodoulopoulos,
  and Mittal}]{thorne-etal-2018-fever}
Thorne, J.; Vlachos, A.; Christodoulopoulos, C.; and Mittal, A.
  2018{\natexlab{b}}.
\newblock {FEVER}: a Large-scale Dataset for Fact Extraction and
  {VER}ification.
\newblock In Walker, M.; Ji, H.; and Stent, A., eds., \emph{Proceedings of the
  2018 Conference of the North {A}merican Chapter of the Association for
  Computational Linguistics: Human Language Technologies}, 809--819.

\bibitem[{Warstadt, Singh, and Bowman(2019)}]{warstadt2019}
Warstadt, A.; Singh, A.; and Bowman, S.~R. 2019.
\newblock Neural Network Acceptability Judgments.
\newblock \emph{Transactions of the Association for Computational Linguistics},
  7: 625--641.

\bibitem[{Wong, Redi, and Saez-Trumper(2021)}]{wong2021}
Wong, K.; Redi, M.; and Saez-Trumper, D. 2021.
\newblock Wiki-Reliability: A Large Scale Dataset for Content Reliability on
  Wikipedia.
\newblock In \emph{Proceedings of the 44th International ACM SIGIR Conference
  on Research and Development in Information Retrieval}, 2437–2442.

\end{thebibliography}

\end{document}